\documentclass{article}

\usepackage[final]{corl_2018} % Uncomment for the camera-ready ``final'' version

\usepackage{graphicx}
\usepackage{caption}
\usepackage{amsmath}
\usepackage{amssymb}
\usepackage{hyperref}
\usepackage{todonotes}
\usepackage{amsmath}
\usepackage[linesnumbered,ruled]{algorithm2e}
\usepackage{tabularx}
\usepackage{bbm}

\usepackage{subcaption} % needed for figures
\usepackage[font=small,labelfont=bf]{caption} % makes figure captions small
\usepackage{wrapfig}

\usepackage[subtle]{savetrees} % attempt to fit more on page

\usepackage[title]{appendix} % appendix

\usepackage{amsmath}

\DeclareMathOperator*{\argmin}{arg\,min}

\title{Dense Object Nets: Learning Dense Visual Object Descriptors By and For Robotic Manipulation}

\author{
  Peter R. Florence*, Lucas Manuelli*, Russ Tedrake\\
  CSAIL, Massachusetts Institute of Technology \\
  \texttt{\{peteflo,manuelli,russt\}@csail.mit.edu} \\
  \textit{*These authors contributed equally to this work.}
}

\begin{document}
\maketitle

%===============================================================================

\begin{abstract}
 
What is the right object representation for manipulation? We would like robots to visually perceive scenes and learn an understanding of the objects in them that (i) is task-agnostic and can be used as a building block for a variety of manipulation tasks, (ii) is generally applicable to both rigid and non-rigid objects, (iii) takes advantage of the strong priors provided by 3D vision, and (iv) is entirely learned from self-supervision.  This is hard to achieve with previous methods: much recent work in grasping does not extend to grasping specific objects or other tasks, whereas task-specific learning may require many trials to generalize well across object configurations or other tasks.  In this paper we present Dense Object Nets, which build on recent developments in self-supervised dense descriptor learning, as a consistent object representation for visual understanding and manipulation. We demonstrate they can be trained quickly (approximately 20 minutes) for a wide variety of previously unseen and potentially non-rigid objects.  We additionally present novel contributions to enable multi-object descriptor learning, and show that by modifying our training procedure, we can either acquire descriptors which generalize across classes of objects, or descriptors that are distinct for each object instance. Finally, we demonstrate the novel application of learned dense descriptors to robotic manipulation. We demonstrate grasping of specific points on an object across potentially deformed object configurations, and demonstrate using class general descriptors to transfer specific grasps across objects in a class. 

\end{abstract}


%===============================================================================

\section{Introduction}

What is the right object representation for manipulation? While task-specific reinforcement learning can achieve impressively dexterous skills for a given specific task \citep{levine2016end}, it is unclear which is the best route to efficiently achieving many different tasks. Other recent work \cite{mahler2017dex,gualtieri2016high} can provide very general grasping functionality but does not address specificity. Achieving specificity, the ability to accomplish specific tasks with specific objects, may require solving the data association problem. At a coarse level the task of identifying and manipulating individual objects can be solved by instance segmentation, as demonstrated in the Amazon Robotics Challenge (ARC) \citep{schwarz2018fast, milan2017semantic} or \citep{jang2017end}. Whole-object-level segmentation, however, does not provide any information on the rich structure of the objects themselves, and hence may not be an appropriate representation for solving more complex tasks. While not previously applied to the robotic manipulation domain, recent work has demonstrated advances in learning dense pixel level data association \citep{choy2016universal,schmidt2017}, including self-supervision from raw RGBD data \citep{schmidt2017}, which inspired our present work.

%Moving beyond visual segmentation, recent work \citep{schmidt2017} marks an advance in learning dense pixel level data association using self-supervision from raw RGBD data.  We have been inspired by \citep{schmidt2017} but we required additional innovations to reliably learn consistent object descriptors and develop a learning method suitable for automation by a robot.
%Additionally, no prior work has: investigated object uniqueness of dense descriptors, demonstrated dense description learning for more than just a single class of objects, or investigated using dense descriptors for manipulation tasks.  In this paper we introduce Dense Object Nets, which are deep neural networks trained to provide dense (pixelwise) description of objects. Dense Object Nets can distinguish multiple objects, be quickly learned in an entirely robotic self-supervised fashion, and enable new manipulation tasks. 

In this paper, we propose and demonstrate using dense visual description as a representation for robotic manipulation.  We demonstrate the first autonomous system that can entirely self-supervise to learn consistent dense visual representations of objects, and ours is the first system we know of that is capable of performing the manipulation demonstrations we provide. Specifically, with no human supervision during training, our system can grasp specific locations on deformable objects, grasp semantically corresponding locations on instances in a class, and grasp specific locations on specific instances in clutter.  Towards this goal, we also provide practical contributions to dense visual descriptor learning with general computer vision applications outside of robotic manipulation.  We call our visual representations Dense Object Nets, which are deep neural networks trained to provide dense (pixelwise) description of objects.%: we show we can learn distinct multi-object descriptors, we can selectively either acquire class-general or instance-specific descriptors,  we demonstrate dense descriptor learning for more than just a single class of objects, and we provide training techniques to improve dense descriptor learning relative to the state of the art.

\textit{\bf{Contributions.}} We believe our largest contribution is that we introduce dense descriptors as a representation useful for robotic manipulation.  %The prior work in dense visual descriptor learning does not involve robots.  
We've also shown that self-supervised dense visual descriptor learning  
can be applied to a wide variety of potentially non-rigid objects and classes (47 objects so far, including 3 distinct classes),
can be learned quickly (approximately 20 minutes), and enables new manipulation tasks.  In example tasks we grasp specific points on objects across potentially deformed configurations, do so with object instance-specificity in clutter, or transfer specific grasps across objects in a class.
We also contribute novel techniques to enable multi-object distinct dense descriptors, and show that by modifying the loss function and sampling procedure, we can either acquire descriptors which generalize across classes of objects, or descriptors that
are distinct for each object instance.
Finally, we contribute general training techniques for dense descriptors which we found
to be critical to achieving good performance in practice.

\textit{\bf{Paper Organization.}}  In Section \ref{sec:related} we describe related work.  As preliminary in Section \ref{pcl-section} we describe the general technique for self-supervising dense visual descriptor learning, which is from \citep{schmidt2017} but reproduced here for clarity.  We then describe additional techniques we've developed for object-centric visual descriptors in Sections \ref{single-object-centric}, and Section \ref{multi-object} describes techniques for distinct multi-object descriptors.   Section \ref{sec:experimental} describes our experimental setup for our autonomous system, and Section 5 describes our results: our learned visual descriptors for a wide variety of objects (Section \ref{single-object-results}) multi-object and selective class generalization (Sections \ref{multi-object-results} and \ref{class-results}), and robotic manipulation demonstrations (Section \ref{grasping-results}).

\section{Related Work}
\label{sec:related}

We review three main areas of related work: learned descriptors, self-supervised visual learning for robots, and robot learning for specific tasks.  In the area of learned visual descriptors our method builds on the approach of Choy et al. \citep{choy2016universal} and Schmidt et al.\ \citep{schmidt2017}.
Other work \citep{thewlis2017unsupervised} uses image warping to learn dense descriptors for a curated dataset of celebrity faces.  None of these prior works in dense visual learning involve robots. 
For cross-instance semantic correspondence, \citep{choy2016universal} relies on human annotations, while \citep{schmidt2017} learns these unsupervised, as we do here.  A thorough comparison to other related work in the area of learned descriptors is provided in \citep{schmidt2017}.
Most work on learning dense correspondences \citep{taylor2012vitruvian,shotton2013scene,brachmann2014learning} requires manually annotated labels. %\citep{simonyan2012descriptor,simo2015discriminative,zagoruyko2015learning} use relative labels to train patch-based, rather than dense, feature descriptors, and rely on global alignment of all known patches.  
Zeng et al. \citep{zeng20173dmatch} also uses dense 3D reconstruction to provide automated labeling, but for descriptors of 3D volume patches. Our work, along with \citep{schmidt2017, thewlis2017unsupervised, brachmann2014learning}, learn descriptors for specific object instances or classes, while \citep{zeng20173dmatch} learns a local volumetric patch descriptor for establishing correspondences between arbitrary partial 3D data.

In the area of self-supervised visual robot learning, while some recent work has sought to understand `how will the world change given the robot's action?'' \citep{finn2017deep,nair2017combining} in this work we instead ask ``what is the current visual state of the robot's world?''. We address this question with a dense description that is consistent across viewpoints, object configurations and (if desired) object classes. At the coarse level of semantic segmentation several works from the Amazon Robotics Challenge used robots to automate the data collection and annotation process through image-level background subtraction \citep{zeng2017robotic,milan2017semantic,schwarz2018fast}. 
In contrast this work uses 3D reconstruction-based change detection and dense pixelwise correspondences, which provides a much richer supervisory signal for use during training. %In addition the background subtraction approach to collecting real labeled data doesn?t extend to the cluttered multi object scenario, whereas our approach is still valid in the multi-object setting.

In the area of robot learning for a specific task there have been impressive works on end-to-end reinforcement learning \citep{levine2016end,zeng2018learning}. In these papers the goal is to learn a specific task, encoded with a reward function, whereas we learn a general task agnostic visual representation. There have also been several works focusing on grasping from RGB or depth images \cite{gualtieri2016high,zeng2017robotic,mahler2017dex,pinto2016supersizing}. These papers focus on successfully grasping any item out of a pile, and are effectively looking for graspable features. They have no consistent object representation or specific location on that object, and thus the robotic manipulation tasks we demonstrate in Section \ref{subsec:robotic_manipulation}, e.g. grasping specific points on an object across potentially deformed object configurations, are out of scope for these works.

\section{Methodology}
\label{sec:methodology}

\subsection{Preliminary: Self-Supervised Pixelwise Contrastive Loss}
\label{pcl-section}

We use self-supervised pixelwise contrastive loss, as developed in \citep{choy2016universal,schmidt2017}. This learns a dense visual descriptor mapping which maps a full-resolution RGB image, $\mathbb{R}^{W \times H \times 3}$ to a dense descriptor space, $\mathbb{R}^{W \times H \times D}$, where for each pixel we have a $D$-dimensional descriptor vector.   Training is performed in a Siamese fashion, where a pair of RGB images, $I_a$ and $I_b$ are sampled from one RGBD video, and many pixel matches and non-matches are generated from the pair of images.  A pixel $u_a \in \mathbb{R}^{2}$ from image $I_a$ is a match with pixel $u_b$ from image $I_b$ if they correspond to the same vertex of the dense 3D reconstruction (Figure \ref{training_procedures} (c-f)).  %taking image $I_a$ at pose $p_a \in SE_3$, raycasting the pixel $u_a$ against the dense reconstruction, projecting into image $I_b$ and appropriately checking for occlusions and field-of-view.\\
The dense descriptor mapping is trained via pixelwise contrastive loss. The loss function aims to minimize the distance between descriptors corresponding to a match, while descriptors corresponding to a non-match should be at least a 
distance $M$ apart, where $M$ is a margin parameter.  The dense descriptor mapping $f(\cdot)$ is used to map an image $I \in \mathbb{R}^{W \times H \times 3}$ to descriptor space $f(I) \in \mathbb{R}^{W \times H \times D}$. Given a pixel 
$u$ we use $f(I)(u)$ to denote the descriptor corresponding to pixel $u$ in image $I$.  We simply round the real-valued pixel $u \in \mathbb{R}^2$ to the closest discrete pixel value $u \in \mathbb{N}^2$, but any continuously-differentiable interpolation can be used for sub-pixel resolution. We denote $D(\cdot)$ as the $L_2$ distance between a pair of pixel descriptors: $D(I_a, u_a, I_b, u_b) \triangleq || f(I_a)(u_a) - f(I_b)(u_b) ||_2$.  At each iteration of training, a large number (on the order of 1 million total) of matches $N_{\text{matches}}$ and non-matches $N_{\text{non-matches}}$ are generated between images $I_a$ and $I_b$. The images are mapped to corresponding descriptor images via $f(\cdot)$ and the loss function is
\begin{align} 
\mathcal{L}_{\text{matches}}(I_a, I_b)  & = \frac{1}{N_{\text{matches}}} \sum_{N_{\text{matches}}} D(I_a, u_a, I_b, u_b)^2 \\
\mathcal{L}_{\text{non-matches}}(I_a, I_b) & = \frac{1}{N_{\text{non-matches}}} \sum_{N_{\text{non-matches}}} max(0, M - D(I_a, u_a, I_b, u_b) )^2 \\
\mathcal{L}(I_a, I_b) & =  \mathcal{L}_{\text{matches}}(I_a, I_b)  +  \mathcal{L}_{\text{non-matches}}(I_a, I_b)
\end{align}
\subsection{Training Procedures for Object-Centric Descriptors}% via Change Detection Masking, Hard-Negative Scaling, Background Randomization, and Data Augmentation}
\label{single-object-centric}

Prior work \citep{schmidt2017} has used dynamic reconstruction \citep{newcombe2015dynamicfusion} of raw RGBD data for only within-scene data association and remarkably showed that even without cross-scene data association, descriptors could be learned that were consistent across many dynamic scenes of the upper body of a human subject. While dynamic reconstruction is powerful, the challenges of topology changes \citep{wei2018} and difficulties of occlusion make it difficult to reliably deploy for an autonomous system.  Schmidt et al.\ \citep{schmidt2017} also used data associations from static scene reconstructions for the task of relocalization in the same static environment.   In contrast we sought to use only static reconstruction but seek consistency for dynamic objects.  Other work \citep{thewlis2017unsupervised} obtains dense descriptor consistency for a curated dataset of celebrity faces using only image warping for data association. %In the human datasets used in \citep{schmidt2017, thewlis2017unsupervised} the human is always facing the camera fairly head on, while we want consistency for all configurations of an object.  

Using our robot mounted camera we are able to reliably collect high quality dense reconstructions for static scenes. Initially we applied only static-scene reconstruction to learn descriptors for specific objects, but we found that the learned object descriptors were not naturally consistent for challenging datasets with objects in significantly different configurations. Subsequently we developed techniques that leverage 3D reconstruction change detection, data augmentation, and loss function balancing to reliably produce consistent object representations with only static-scene data association for the wide variety of objects we have tested.  These techniques also improve the precision of correspondences, as is discussed in Section \ref{single-object-results}.  While we have tried many other ideas, these are the techniques that were empirically found to significantly improve performance. 
\begin{center}
  \includegraphics[keepaspectratio=true,scale=0.39]{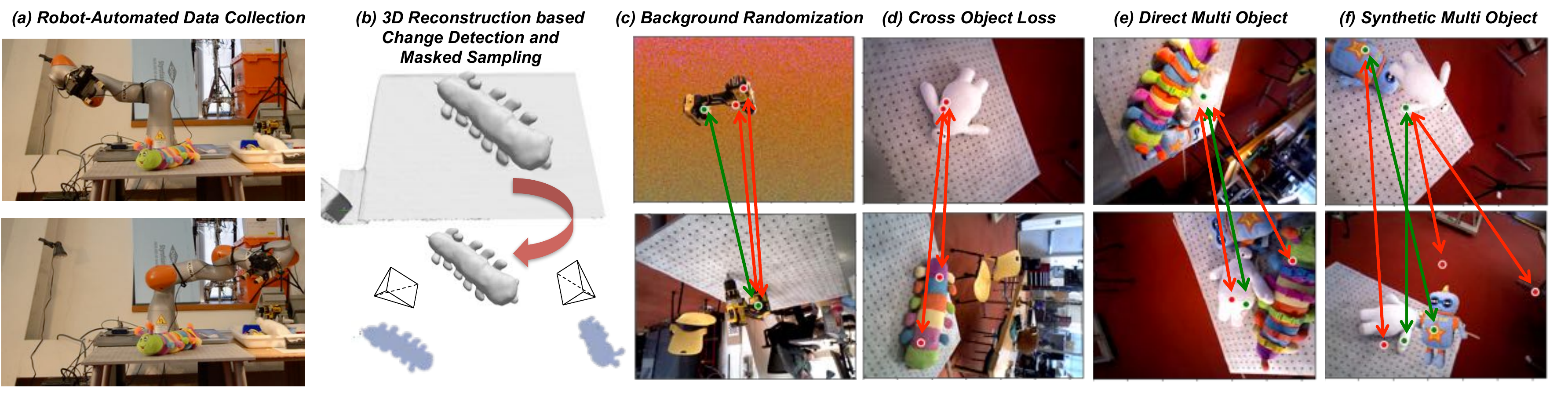}
  \captionof{figure}{Overview of the data collection and training procedure. (a) automated collection with a robot arm. (b) change detection using the dense 3D reconstruction. (c)-(f) matches depicted in green, non-matches depicted in red. }\label{training_procedures}
\end{center}

\textit{\bf{Object masking via 3D change detection.}}  %We would like to learn object-centric descriptors, but do so in a way that is learnable without special (i.e. green-background) environments.  
Since we are trying to learn descriptors of objects that take up only a fraction of a full image, we observe significant improvements if the representational power of the models are focused on the objects rather than the backgrounds. A $640 \times 480$ image
contains $307,200$ pixels but an image in our dataset may have as few as 1,000 to 10,000 of those pixels, or .3\%-3\%, that correspond to the object of interest. 
Initial testing with human-labeled object masks \citep{marion2017pipeline} showed that if matches for data associations were sampled only on the object (while non-matches were sampled from the full image) then correspondence performance was significantly improved.  In order to provide autonomous object masking without any human input, we leverage our 3D reconstructions and results from the literature on 3D change detection \citep{finman2013toward} to recover the object-only part of the reconstruction (Figure \ref{training_procedures}b).  Projecting this geometry into each camera frame yields object masks for each image. 
% While we have implemented general 3D-differencing that can allow object masking in even cluttered environments, for our purposes of object manipulation on top of a flat table, we can simply threshold above the table to recover the object-only part of the reconstruction.
We want to emphasize that automatic object masking enables many other techniques in this paper, including: background domain randomization, cross-object loss, and synthetic multi-object scenes.  

%We emphasize that no prior knowledge is needed: the meshes are only partial meshes (there is no data about unseen parts of the object) and are not currently used as any global representation of the object.

%We discuss non-match sampling in supplementary material.

\textit{\bf{Background domain randomization.}} A strategy to encourage cross-scene consistency is to enforce that the learned descriptors are not reliant on the background.  Since we have autonomously acquired object masks, we can  domain randomize \citep{tobin2017domain} the background (Figure \ref{training_procedures}c top) to encourage consistency -- rather than memorizing the background (i.e. by describing the object by where it is relative to a table edge), the descriptors are forced to be representative of only the object.  

\textit{\bf{Hard-negative scaling.}} Although as in \citep{schmidt2017} we originally normalized $\mathcal{L}_{\text{matches}}$ and $\mathcal{L}_{\text{non-matches}}$ by $N_{\text{matches}}$ and $N_{\text{non-matches}}$, respectively, we found that what we call the ``hard-negative rate'', i.e. the percentage of sampled non-matches for which $M - D(I_a, u_a, I_b, u_b) > 0$ would quickly drop well below 1\% during training.  While not precisely hard-negative mining \citep{felzenszwalb2010object}, we empirically measure improved performance if rather than scaling  $\mathcal{L}_{\text{non-matches}}$ by $N_{\text{non-matches}}$, we adaptively scale by the number of hard negatives in the non-match sampling, $N_{\text{hard-negatives}}$, where $\mathbbm{1}$ is the indicator function:
\begin{align} 
N_{\text{hard-negatives}} & =  \sum_{N_{\text{non-matches}}} \mathbbm{1}(M - D(I_a, u_a, I_b, u_b) > 0) \\
\mathcal{L}_{\text{non-matches}}(I_a, I_b) & = \frac{1}{N_{\text{hard-negatives}}} \sum_{N_{\text{non-matches}}} max(0, M - D(I_a, u_a, I_b, u_b) )^2 
\end{align}
\textit{\bf{Data diversification and augmentation.}} While we collect only a modest number of scenes (4-10) per object or class, we ensure they are diverse.  We physically acquire a diverse set of orientations, crops, and lighting conditions by utilizing the full workspace of our 7-DOF robot arm, randomizing end-effector rotations with an approximate gaze constraint towards the collection table, and varying lighting conditions. We also applied synthetic 180-degree rotations randomly to our images.

\subsection{Multi-Object Dense Descriptors}
\label{multi-object}

We of course would like robots to have dense visual models of more than just one object.  When we began this work it wasn't obvious to us what scale of changes to our training procedure or model architecture would be required in order to simultaneously (a) achieve individual single-object performance comparable to a single-object-only model, while also (b) learn dense visual descriptors for objects that are \textit{globally distinct} -- i.e., the bill of a hat would occupy a different place in descriptor space than the handle of a mug.   
To achieve distinctness, we introduce three strategies: 

\textit{\bf{i. Cross-object loss.}} The most direct way to ensure that different objects occupy different subsets of descriptor space is to directly impose \textit{cross-object loss} (Figure \ref{training_procedures}d).  Between two different objects, we know that each and every pair of pixels between them is a non-match.  Accordingly we randomly select two images of two different objects, randomly sample many pixels from each object (enabled by object masking), and apply non-match loss (with hard-negative scaling) to all of these pixel pairs.

\textit{\bf{ii. Direct training on multi-object scenes.}} A nice property of pixelwise contrastive loss, with data associations provided by 3D geometry, is that we can directly train on multi-object, cluttered scenes \textit{without any individual object masks} (Figure \ref{training_procedures}e).  This is in contrast with training pixelwise semantic segmentation, which requires labels for each individual object in clutter that may be difficult to attain, i.e. through human labeling.  With pixel-level data associations provided instead by 3D geometry, the sampling of matches and the loss function still makes sense, even in clutter.

\textit{\bf{iii. Synthetic multi-object scenes.}} We can also synthetically create multi-object scenes by layering object masks  \cite{schwarz2018fast}. To use dense data associations through synthetic image merging, we prune matches that become occluded during layering (Figure \ref{training_procedures}f).  A benefit of this procedure is that we can create a combinatorial number of ``multi-object'' scenes from only single object-scenes, and can cover a wide range of occlusion types without collecting physical scenes for each.  %At this time we have made no effort to ensure ``realism'' of multi-object scenes -- although this could be future work. %It is unclear if photo-realism of the synthetic multi object scenes is important for ultimate performance.

\section{Experimental}
\label{sec:experimental}

\textit{\bf{Data Collection and Pre-Processing.}} The minimum requirement for raw data is to collect an RGBD video of an object or objects. Figure \ref{training_procedures} shows our experimental setup; we utilize a 7-DOF robot arm (Kuka IIWA LBR) with an RGBD sensor (Primesense Carmine 1.09) mounted at the end-effector.  With the robot arm, data collection can be highly automated, and we can achieve reliable camera poses by using forward kinematics along with knowledge of the camera extrinsic calibration.  For dense reconstruction we use TSDF fusion \citep{curless1996volumetric} of the depth images with camera poses provided by forward kinematics. 
An alternative route to collecting data which does not require a calibrated robot is to use a dense SLAM method (for example, \citep{newcombe2011kinectfusion,whelan2015elasticfusion}). 
In between collecting RGBD videos, the object of interest should be moved to a variety of configurations, and the lighting can be changed if desired.  While for many of our data collections a human moved the object between configurations, we have also implemented and demonstrated (see our video) the robot autonomously rearranging the objects, which highly automates the object learning process.  We employ a Schunk two-finger gripper and plan grasps directly on the object point cloud (Appendix \ref{appendix:grasping}).  If multiple different objects are used, currently the human must still switch the objects for the robot and indicate which scenes correspond to which object, but even this information could be automated by the robot picking objects from an auxiliary bin. % and use continuity to simply identify which logs are of the same object.

\textit{\bf{Training Dense Descriptors.}} For training, at each iteration we randomly sample between some subset of specified image comparison types (Single Object Within Scene, Different Object Across Scene, Multi Object Within Scene, Synthetic Multi Object), 
and then sample some set of matches and non-matches for each.  In this work, we use only static-scene reconstructions, so pixel matches between images can be easily found by raycasting and reprojecting against the dense 3D reconstruction model, and appropriately checking for occlusions and field-of-view constraints. 
For the dense descriptor mapping we train a 34-layer, stride-8 ResNet pretrained on ImageNet, but we expect any fully-convolutional network (FCN) that has shown effectiveness on semantic segmentation tasks to work well.
Additional training details are contained in the Appendix \ref{appendix:training}.  
%For within-scene image comparison types, example standard parameters are sampling 10,000 pixels off the object mask in one image %, attempting to find matches for each of these samples in the other image, 
%and then sampling 150 non-matches for each valid match.  We have experimented with various non-match sampling strategies, including biasing non-match sampling to also be masked on the object, but none of these strategies have proven to be measurably more effective than uniform random sampling non-match pixels from the full image.    Networks are trained with an Adam optimizer and we often achieve reasonable performance within 3500 steps.

\section{Results}\label{results}

\begin{center}
  \includegraphics[keepaspectratio=true,scale=0.25]{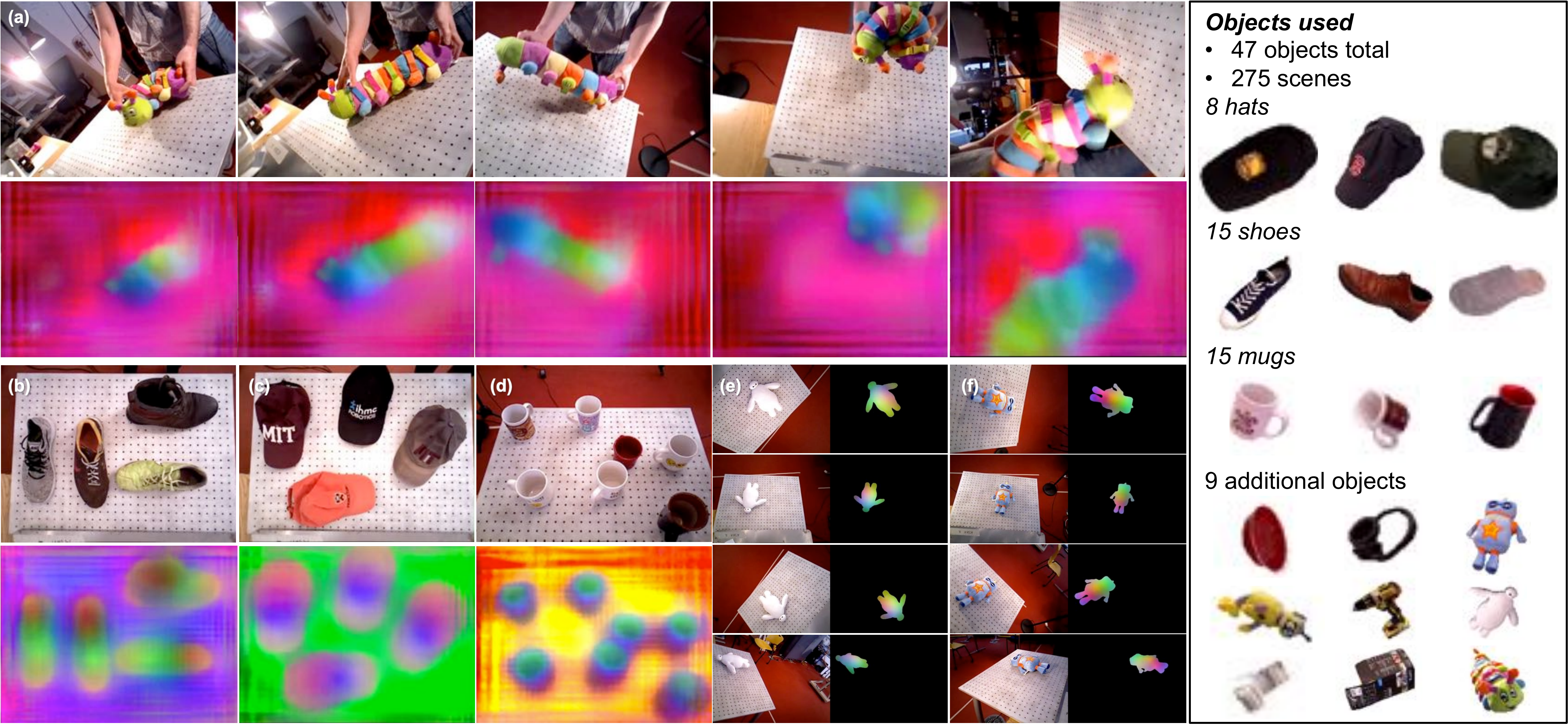}
  \captionof{figure}{Learned object descriptors can be consistent across significant deformation (a) and, if desired, across object classes (b-d). Shown for each (a) and (b-d) are RGB frames (top) and corresponding descriptor images (bottom) that are the direct output of a feed-forward pass through a trained network.
 (e)-(f) shows that we can learn descriptors for low texture objects, with the descriptors masked for clear visualization.  Our object set is also summarized (right).}\label{descriptors_variety}
\end{center}

\subsection{Single-Object Dense Descriptors}\label{single-object-results}

\begin{figure}
	\centering
  \includegraphics[keepaspectratio=true,width=\textwidth]{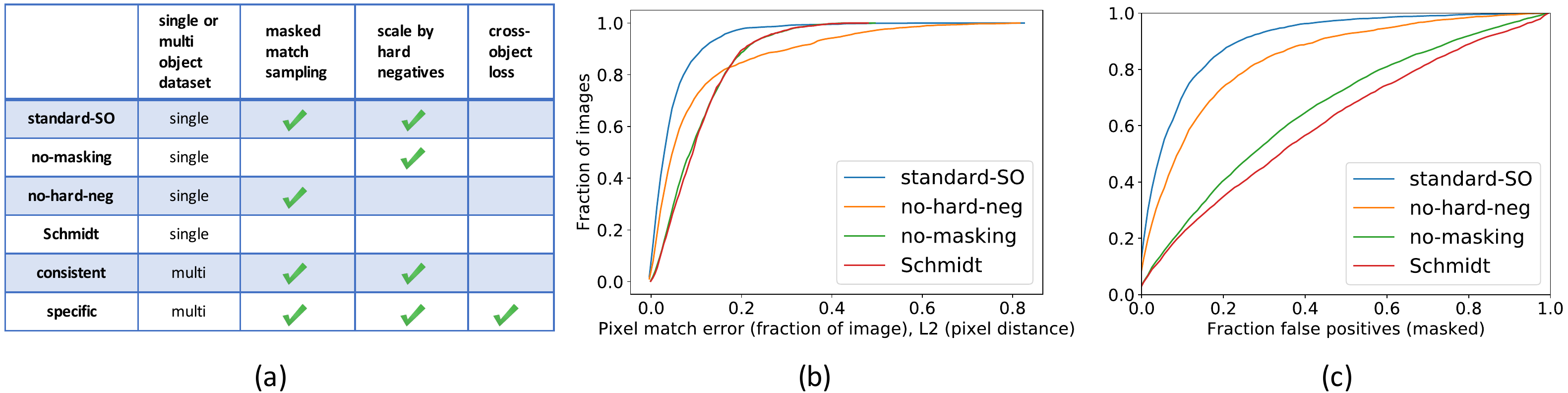}
  \captionof{figure}{(a) table describing the different types of networks referenced in experiments. Column labels correspond to techniques described in Section \ref{sec:methodology}. (a) Plots the cdf of the L2 pixel distance (normalized by image diagonal, 800 for a 640 x 480 image) between the best match $\hat{u}_b$ and the true match $u_b^*$, e.g. for \textbf{standard-SO} in $93\%$ of image pairs the normalized pixel distance between $u_b^*$ and $\hat{u}_b$ is less than $13\%$. All networks were trained on the same dataset using the labeled training procedure from (a). (c) Plots the cdf of the fraction of pixels $u_b$ of the object pixels with $D(I_a, u_a^*, I_b, u_b) < D(I_a, u_a^*, I_b, u_b^*)$, i.e. they are closer in descriptor space to $u_a^*$ than the true match $u_b^*$.}\label{fig:network_types}
\end{figure}

\begin{figure}
	\begin{center}
    \includegraphics[width=0.70\textwidth]{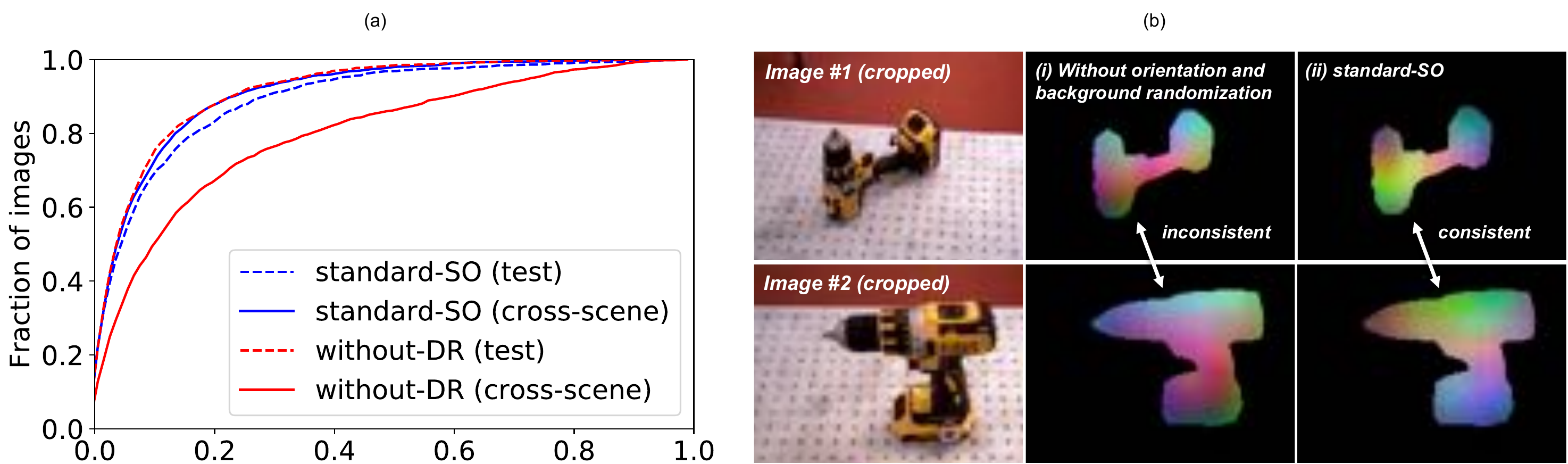}
    \end{center}
  \caption{ (a), with same axes as Figure \ref{fig:network_types}a, compares \textbf{standard-SO} with \textbf{without-DR}, for which the only difference is that \textbf{without-DR} used no background domain randomization during training.  The dataset used for (a) is of three objects, 4 scenes each.  (b) shows that for a dataset containing 10 scenes of a drill, learned descriptors are inconsistent without background and orientation randomization during training (middle), but consistent with them (right). }
  \label{fig:data_augmentation}
\end{figure}

We observe that with our training procedures described in Section \ref{single-object-centric}, for a wide variety of objects we can acquire dense descriptors that are consistent across viewpoints and configurations.  The variety of objects includes moderately deformable objects such as soft plush toys, shoes, mugs, and hats, and can include very low-texture objects (Figure \ref{descriptors_variety}). Many of these objects were just grabbed from around the lab (including the authors' and labmates' shoes and hats), and dense visual models can be reliably trained with the same network architecture and training parameters. The techniques in Section \ref{single-object-centric} provide significant improvement in both (a) qualitative consistency over a wide variety of viewpoints, and (b) quantitative precision in correspondences.  As with other works that learn pairwise mappings to some descriptor space \citep{schroff2015facenet}, in practice performance can widely vary based on specific sampling of data associations and non-associations used during training. One way to quantitatively evaluate correspondence precision is with human-labeled (used only for evaluation; never for training) correspondences across two images of an object in \textit{different} configurations. Given two images $I_a,I_b$ containing the same object and pixel locations $u_a^* \in I_a, u_b^* \in I_b$ corresponding to the same physical point on the object, we can use our dense descriptors to estimate $u_b^*$ as $\hat{u}_b$:
\begin{equation}
\hat{u}_b \triangleq \argmin_{u_b \in I_b} D(I_a, u_a^*, I_b, u_b)
\label{eq:best_match}
\end{equation}
Figure \ref{fig:network_types} (b-c) shows a quantitative comparison of ablative experiments, for four different training procedures \textbf{standard-SO}, \textbf{no-hard-neg}, \textbf{no-masking} and \textbf{Schmidt} (described in Figure \ref{fig:network_types}a). \textbf{standard-SO} performs significantly better than \textbf{Schmidt}, and the performance of \textbf{standard-SO} compared to \textbf{no-hard-neg}, \textbf{no-masking} shows that both object masking and hard-negative scaling are important for good performance.
We also find that for some low-texture objects, orientation randomization and background domain randomization are critical for attaining consistent object descriptors. Otherwise the model may learn to memorize which side of the object is closest the table, rather than a consistent object model (Figure \ref{fig:data_augmentation}c). Background domain randomization is most beneficial for smaller datasets, where it can significantly reduce overfitting and encourage consistency (Figure \ref{fig:data_augmentation}b); it is less critical for high-texture objects and larger datasets.

\begin{figure}
	\centering
	\begin{subfigure}[b]{1.0\textwidth}
        \includegraphics[width=\textwidth]{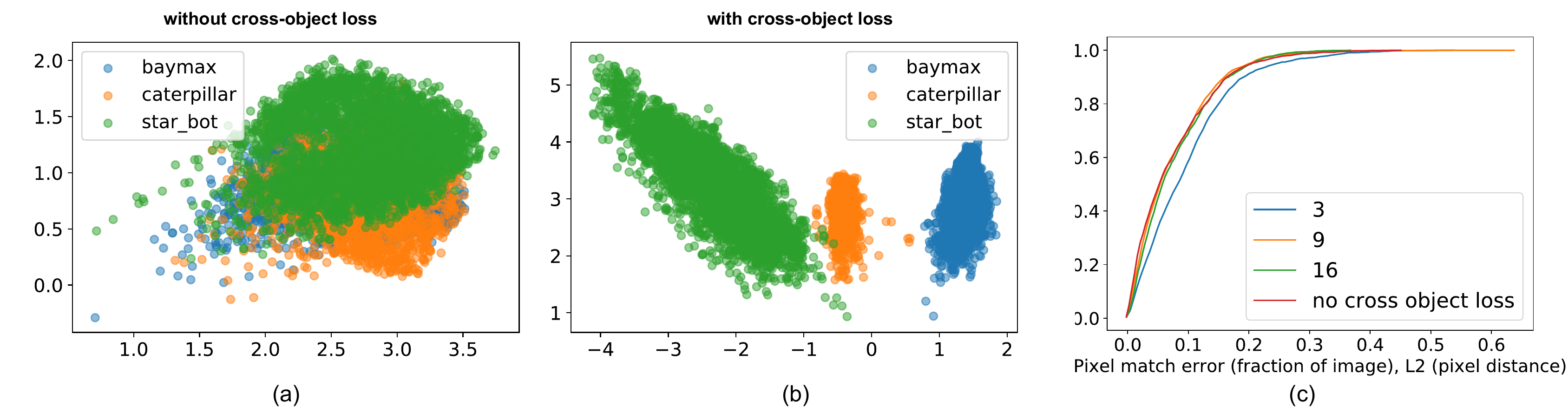}
    \end{subfigure}
    \caption{Comparison of training without any distinct object loss (a)  vs. using cross-object loss (b).  In (b), 50\% of training iterations applied cross-object loss and 50\% applied single-object within-scene loss, whereas (a) is 100\% single-object within-scene loss.  The plots show a scatter of the descriptors for 10,000 randomly-selected pixels for each of three distinct objects.  Networks were trained with $D=2$ to allow direct cluster visualization. (c) Same axes as Figure \ref{fig:network_types} (a). All networks were trained on the same 3 object dataset. Networks with a number label were trained with cross object loss and the number denotes the descriptor dimension. no-cross-object is a network trained without cross object loss. }
    \label{fig:distinct_objects}
\end{figure}

\subsection{Multi-Object Dense Descriptors}\label{multi-object-results}

An early observation during experimentation was that \textit{overlap} in descriptor space naturally occurs if the same model is trained simultaneously on different singulated objects, where sampling of matches and non-matches was only performed \textit{within scene}.  Since there is no component of the loss function that requires different objects to occupy different subsets of descriptor space, the model maps them to an overlapping subset of descriptor space, distinct from the background but not each other (Figure \ref{fig:distinct_objects}a).  Accordingly we sought to answer the question of whether or not we could \textit{separate} these objects into unique parts of descriptor space.  

By applying cross-object loss (Section \ref{multi-object}.i, training mode \textbf{specific} in Figure \ref{fig:network_types}a), we can convincingly separate multiple objects such that they each occupy distinct subsets of descriptor space (Figure \ref{fig:distinct_objects}). Note that cross-object loss is an extension of sampling \textit{across scene} as opposed to only \textit{within scene}.
Given that we can separate objects in descriptor space, 
we next investigate: 
does the introduction of object distinctness significantly limit the ability of the models to achieve correspondence precision for each individual object?  
For multi-object datasets, we observe that there is a measurable decrease in correspondence precision for small-dimensional descriptor spaces when the cross-object loss is introduced, but we can recover correspondence precision by training slightly larger-dimensional descriptor spaces (Figure \ref{fig:distinct_objects}c). For the most part, 3-dimensional descriptor spaces were sufficient to achieve saturated (did not improve with higher-dimension) correspondence precision for single objects, yet this is often not the case for distinct multi-object networks.  %For example a 3-dimensional descriptor network trained without cross-object loss can achieve good correspondence precision on a few different objects, this precision is measurably lower when cross-object loss is introduced, and the precision is approximately recovered by increasing to for example a 9-dimensional descriptor network.

\subsection{Selective Class Generalization or Instance Specificity}\label{class-results}

Surprisingly we find that when trained simultaneously on similar items of a class using training mode \textbf{consistent}, the learned descriptors naturally generalize well across sufficiently similar instances of the class. This result of converging descriptors across a class is similar to the surprising generalization observed for human datasets in \citep{schmidt2017, thewlis2017unsupervised}. Here we show that we can obtain class consistent dense descriptors for 3 different classes of objects (hats, shoes, and mugs) trained with only static-scene data association.
We observe that the descriptors are consistent despite considerable differences in color, texture, deformation, and even to some extent underlying shape. 
The training requirements are reasonably modest -- only 6 instances of hats were used for training yet the descriptors generalize well to unseen hats, including a blue hat, a color never observed during training.  
The generalization extends to instances that a priori we thought would be failure modes: we expected the boot (Figure \ref{all_picks}h) to be a failure mode but there is still reasonable consistency with other shoes.  Sufficiently different objects are not well generalized, however -- for example Baymax and Starbot (Figure \ref{descriptors_variety}e,f) are both anthropomorphic toys but we do not attain general descriptors for them.  
While initially we proposed research into further encouraging consistency within classes, for example by training a Dense Object Net to fool an instance-discriminator, the level of consistency that naturally emerges is remarkable and was sufficient for our desired levels of precision and applications.

For other applications, however, instance-specificity is desired. 
For example, what if you would like your robot to recognize a certain point on hat A as distinct from the comparable point on hat B?  Although we could separate very distinct objects in multi-object settings as discussed in the previous section, it wasn't obvious to us if we could satisfactorily separate objects of the same class.  We observe, however, that by applying the multi-object techniques (\textbf{specific} in Figure \ref{fig:network_types}) previously discussed, we can indeed learn distinct descriptors even for very similar objects in a class (Figure \ref{all_picks}iv).

\subsection{Example Applications to Robotic Manipulation: Grasping Specific Points}\label{grasping-results}
\label{subsec:robotic_manipulation}

%What motivated us to learn dense visual descriptors of objects was the application to robotic manipulation.  
%With good visual models of object-specific descriptors, we believe there are great possibilities for robots understanding arbitrary objects in a detailed way, and using these models to effectively manipulate objects.  
Here we demonstrate a variety of manipulation applications in grasping specific points on objects, where the point of interest is specified in a reference image.  We emphasize there could be many other applications, as mentioned in the Conclusion. In our demonstrations, a user clicks on just one pixel $u_a^*$ in one reference image. Now the robot has the ability to autonomously identify the corresponding point in new scenes via Equation \ref{eq:best_match}.
%
%from many different configurations of both the object and the camera. Given a set of images $I_0, ..., I_{N-1}$ the best match for the reference point is simply the pixel $u$ and image index $i$ that minimizes %$\underset{i, u}{min} 
%$||f(I_i)(u) - d^*||$.
%
Akin to other works with similarity learning in metric spaces \citep{schroff2015facenet}, we set a simple threshold to determine whether a valid match exists. If a match is identified in the new scene we can instruct the robot to autonomously grasp this point by looking up the corresponding location in the point cloud and using simple geometric grasping techniques (details in Appendix \ref{appendix:grasping}).

The particular novel components of these manipulation demonstrations are in grasping the visual corresponding points for arbitrary pixels that are either in different (potentially deformed) configurations (Fig. \ref{all_picks}i-ii), general across instances of classes (Fig. \ref{all_picks}iii), or instance-specific in clutter (Fig. \ref{all_picks}iv). Our video best displays these tasks.\footnote{See video (\url{https://youtu.be/L5UW1VapKNE}) for extensive videos of the different types of robot picking.} Note that only a dense (as opposed to sparse) method can easily accommodate the arbitrary selection of interaction points, and class-generalization is out of scope for hand-designed descriptors such as SIFT. This is also out of scope for general grasp planners like \cite{gualtieri2016high,zeng2017robotic,mahler2017dex,pinto2016supersizing} which lack any visual object representation, and for segmentation based methods \citep{zeng2017robotic,milan2017semantic,schwarz2018fast} since the visual representation provided by segmentation doesn't capture any information beyond the object mask.

\begin{center}
  \includegraphics[keepaspectratio=true,scale=0.42]{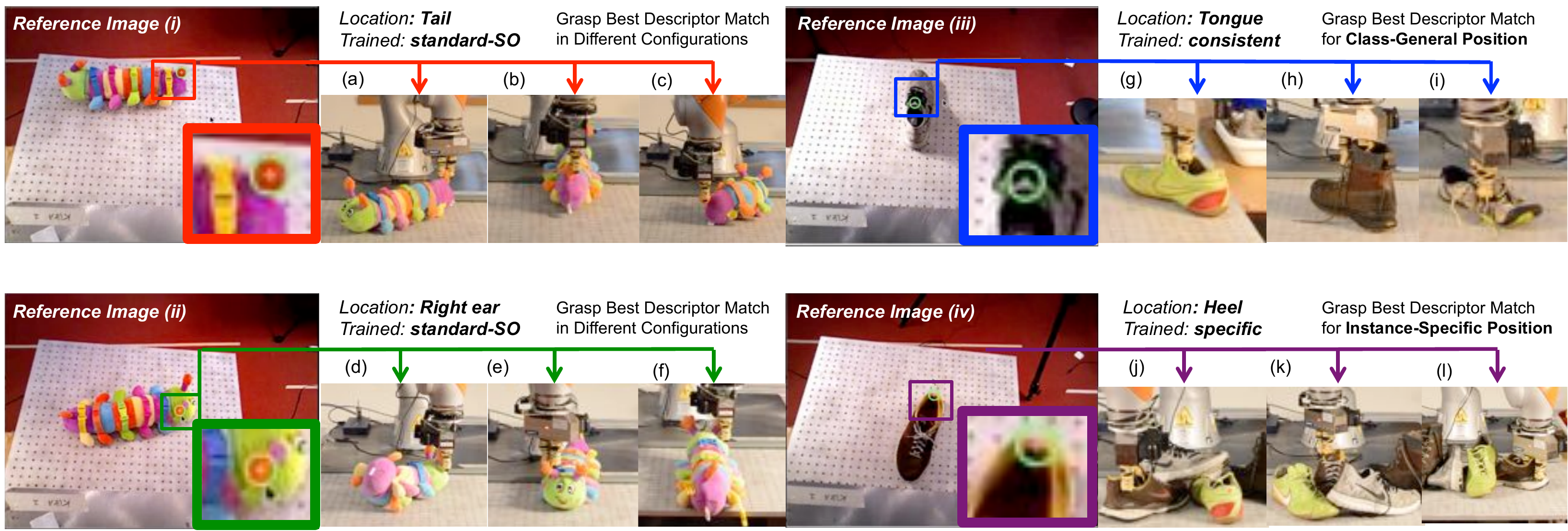}
  \captionof{figure}{Depiction of ``grasp specific point'' demonstrations.  For each the user specifies a pixel in a single reference image, and the robot automatically grasps the best match in test configurations.  For single-object, two different points for the caterpillar object are shown: tail (i) and right ear (ii). Note that the ``right-ear'' demonstration is an example of the ability to break symmetry on reasonably symmetrical objects. %Given that the right and left ear of the caterpillar look very similar a small patch based method would probably struggle in this situation, whereas the receptive field of our network architecture is large. 
For class generalization (iii), trained with \textbf{consistent}, the robot grasps the class-general point on a variety of instances.  This was trained on only 4 shoes and extends to unseen instances of the shoe class, for example (c).  For instance-specificity (iv) trained with \textbf{specific} and augmented with synthetic multi object scenes (\ref{multi-object}.iii), the robot grasps this point on the specific instance even in clutter.}\label{all_picks}
\end{center}

\section{Conclusion}

This work introduces Dense Object Nets as visual object representations which are useful for robotic manipulation and can be acquired with only robot self-supervision. Building on prior work on learning pixel-level data associations we develop new techniques for object-centricness,
multi-object distinct descriptors, and learning dense descriptors \textit{by and for} robotic manipulation. Without these object centric techniques we found that data associations from static-scene reconstructions were not sufficient to achieve consistent object descriptors. Our approach has enabled automated and reliable descriptor learning at scale for a wide variety of objects (47 objects, and 3 classes).  
We also show how learned dense descriptors can be extended to the multi object setting. With new contrastive techniques we are able to train Dense Object Nets that map different objects to different parts of descriptor space. 
Quantitative experiments show we can train these multi object networks while still retaining the performance of networks that do not distinguish objects. We also can learn class-general descriptors which generalize across different object instances, and demonstrated this result for three classes: shoes, hats, and mugs. Using class-general descriptors we demonstrate a robot transferring grasps across different instances of a class. Finally we demonstrate that our distinct-object techniques work even for objects which belong to the same class. This is demonstrated by the robot grasping a specific point on a target shoe in a cluttered pile of shoes.
We believe Dense Object Nets can enable many new approaches to robotic manipulation, and are a novel object representation that addresses goals (i-iv) stated in the abstract. %Dense Object Nets are (i) task-agnostic and can be used as a building block for a variety of manipulation tasks, (ii) generally applicable to both rigid and non-rigid objects, (iii) take advantage of the strong priors provided by 3D vision, and (iv) are entirely learned from self-supervision. 
In future work we are interested to explore new approaches to solving manipulation problems that exploit the dense visual information that learned dense descriptors provide, and how these dense descriptors can benefit other types of robot learning, e.g. learning how to grasp, manipulate and place a set of objects of interest. 

%===============================================================================

% The maximum paper length is 8 pages excluding references and acknowledgements, and 10 pages including references and acknowledgements

\clearpage
% The acknowledgments are automatically included only in the final version of the paper.
\acknowledgments{ The authors thank
Duy-Nguyen Ta (calibration), Alex Alspach (hardware), Yunzhi Li (training techniques), Greg Izatt (robot software), and Pat Marion (perception and visualization) for their help.  We also thank Tanner Schmidt for helpful comments in preparing the paper.
This work was supported by:
Army Research Office, Sponsor Award No. W911NF-15-1-0166; 
Draper Laboratory Incorporated, Sponsor Award No. SC001-0000001002;
Lincoln Laboratory/Air Force, Sponsor Award No. 7000374874;
Amazon Research Award, 2D-01029900;
Lockheed Martin Corporation, Sponsor Award No. RPP2016-002.
Views expressed in the paper are not endorsed by the sponsors.}

%===============================================================================

\bibliography{corl_2018}  % .bib

\newpage

\begin{appendices}
\section{Experimental Hardware}
\label{appendix:hardware}
\begin{figure}
    \centering
    \begin{subfigure}[b]{0.45\textwidth}
        \includegraphics[width=\textwidth]{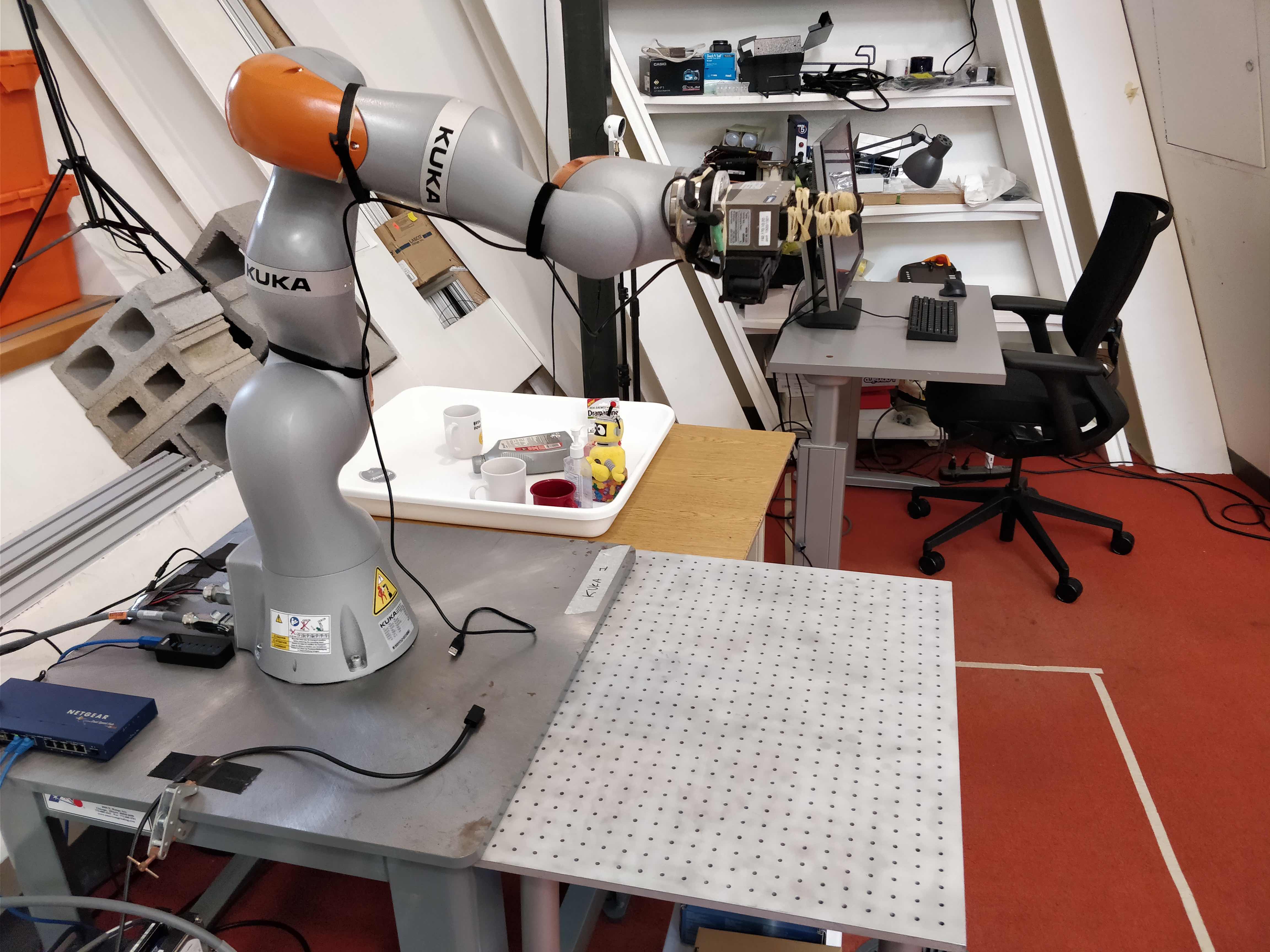}
        \caption{}
        \label{fig:tricks_pixel_match_error}
    \end{subfigure}
    \centering
    \begin{subfigure}[b]{0.45\textwidth}
        \includegraphics[width=\textwidth]{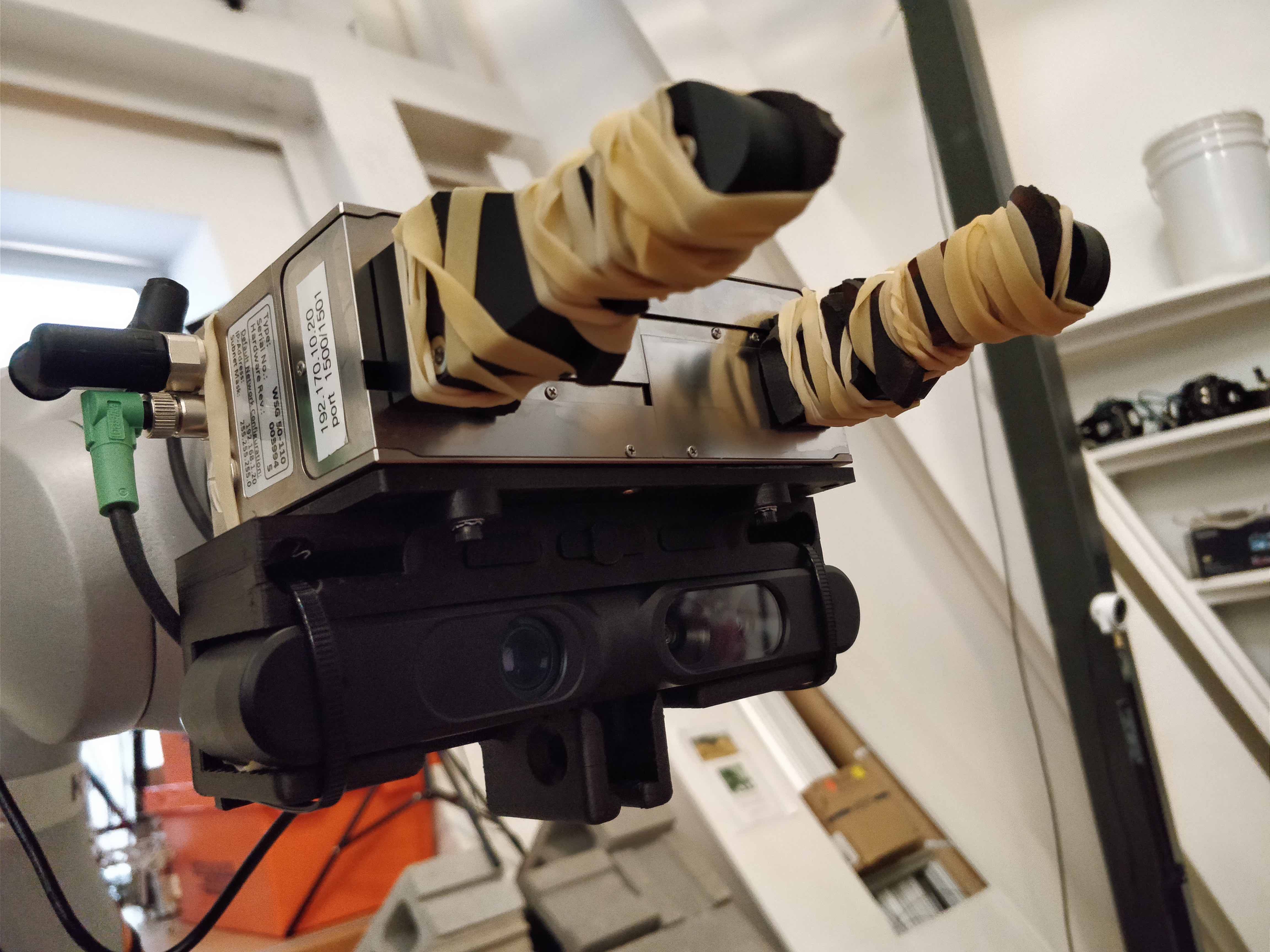}
        \caption{}
        \label{fig:tricks_fraction_false_positives}
    \end{subfigure}
    \caption{(a) Kuka IIWA LRB robot arm. (b) Schunk WSG 50 gripper with Primesense Carmine 1.09 attached}
    \label{fig:kuka_hardware} 
\end{figure}

All of our data was collected using an RGBD camera mounted on the end effector of a 7 DOF robot arm (see Figure \ref{fig:kuka_hardware}). We used a Kuka IIWA LBR robot with a Schunk WSG 50 parallel jaw gripper. A Primesense Carmine 1.09 RGBD sensor was mounted to the Schunk gripper and precisely calibrated for both intrinsics and extrinsics.

\section{Experimental Setup: Data Collection and Pre-Processing}
\label{appendix:experimental_setup}
As discussed in the paper all of our data consists of 3D dense reconstructions of a static scene. To collect data for a single scene we place an object (or set of objects) on a table in front of the robot. We then perform a scanning pattern with the robot which allows the camera to capture many different viewpoints of the static scene. This procedure takes about 70 seconds during which approximately 2100 RGBD frames are captured (the sensor outputs at 30Hz). Using the forward kinematics of the robot, together with our camera extrinsic calibration, we also record the precise camera pose corresponding to each image. Since we have the camera poses corresponding to each depth image we use TSDF fusion \citep{curless1996volumetric} to obtain a dense 3D reconstruction. Although we use our robot's forward kinematics to produce the dense 3D reconstruction together with camera pose tracking, any dense SLAM method (such as \citep{newcombe2011kinectfusion}) could be used instead. In practice we found that using the robot's forward kinematics to obtain camera pose estimates produces very reliable 3D reconstructions which are robust to lack of geometric or RGB texture, varying lighting conditions, etc. Next we obtain new depth images for each recorded camera frame by rendering against the 3D reconstruction using our camera pose estimates. This step produces depth images which are globally consistent and free of artifacts (i.e. missing depth values, noise etc.). This keeps the process of finding correspondences across RGB images as as a simple operation between poses and depth images. To enable the specific training techniques discussed in the paper we also need to know which parts of the scene correspond to the objects of interest. To do this we implemented the change detection technique of \citep{finman2013toward}. In practice since all of our data was collected on a tabletop, and our reconstructions can be easily globally aligned (due to the fact that we know the global camera poses from the robot's forward kinematics) we can simply crop the reconstruction to the area above the table. Once we have isolated the part of the reconstruction corresponding to the objects of interest, we can easily render binary masks via the same procedure as was used to generate the depth images.

Our RGBD sensor captures images at 30Hz, so we downsample the images to avoid having images which are too visually similar. Specifically we downsample images so that the camera poses are sufficiently different (at least 5cm of translation, or 10 degrees of rotation). After downsampling we are left with approximately 315 images per scene.

In between collecting RGBD videos, the object of interest should be moved to a variety of configurations, and the lighting can be changed if desired.  While for many of our data collections a human moved the object between configurations, we have also implemented and demonstrated (see our video) the robot autonomously rearranging the objects, which highly automates the object learning process.  We employ a Schunk two-finger gripper and plan grasps directly on the object point cloud (see Appendix \ref{appendix:grasping}).  If multiple different objects are used, currently the human must still switch the objects for the robot and indicate which scenes correspond to which object, but even this information could be automated by the robot picking objects from an auxiliary bin and use continuity to simply identify which logs are of the same object.

\section{Grasping Pipeline}
\label{appendix:grasping}
While our learned visual descriptors can help us determine \textit{where} to grasp, they can't be used during the bootstrapping phase before visual learning has occurred, and they don't constrain the 6DOF orientation of the gripper.  Accordingly to choose grasps our pipeline employs simple geometric point cloud based techniques.   There are two types of grasping that are performed in the paper. The first is performed while grasping the object to automatically reorient it during data collection, during the visual learning process. To achieve this we first use the depth images and camera poses to fuse a point cloud of the scene.  We then randomly sample many grasps on the point cloud and prune those that are in collision. The remaining collision free grasps are then scored by a simple geometric criterion that evaluates the grasps for antipodality. The highest scoring grasp is then chosen for execution.  The second type of grasping is, as in Section 5.4 of the paper, when we are attempting to grasp a specific point.  In this case the robot moves to a handful of pre-specified poses and records those RGBD images.  The RGB images are used to look up the best descriptor match and determine a pixel space location to grasp in one image, and the corresponding depth image and camera pose are used to determine the 3D location of this point.  The handful of depth images are also fused into a point cloud, and the grasping procedure is almost the same as the first type, with the slight modification that all the randomly sampled grasps are centered around the target grasp point. Although there are a variety of learning based grasping algorithms \citep{gualtieri2016high, mahler2017dex} that could have been used, we found that our simple geometric based grasp planning was sufficient for the tasks at hand.

\section{Network Architecture and Training Details}
\label{appendix:training}
For our network we use 34-layer, stride-8 ResNet (pretrained on ImageNet), and then bilinearly upsample to produce a full resolution $640 x 480$ image.

For training, at each iteration we randomly sample between some specified subset of specified image comparison types (Single Object Within Scene, Different Object Across Scene, Multi Object Within Scene, Synthetic Multi Object).  The weighting between how often each type is chosen is done via specifying their probabilities of being selected.  Once the type has been sampled we then sample some set of matches and non-matches for each (around 1 million in total). Each step of the optimizer uses a single pair of images.

All the networks were trained with the same optimizer settings. Networks were trained for 3500 steps using an Adam optimizer with a weight decay of $1e-4$. Training was performed on a single Nvidia 1080 Ti or Titan Xp, a single step takes approximately $0.22$ seconds, i.e. approximately 13 minutes, and so together with collecting a handful of scenes the entire training for a new object can take 20 minutes. The learning rate was set to $1e-4$ and dropped by $0.9$ every $250$ iterations. The networks trained with procedure $\textbf{specific}$ used a 50-50 split of within scene image pairs and across scene image pairs $50\%$ of the time. For the network used to grasp the heel of the red/brown shoe in Section 5.4 we sampled equally the three data types (Single Object Within Scene, Different Object Across Scene, Synthetic Multi Object).

\end{appendices}

\end{document}